\documentclass[letterpaper, 10 pt, conference]{ieeeconf}  

\IEEEoverridecommandlockouts                              

\usepackage{graphics} 
\usepackage{times} 
\usepackage{amsmath} 
\usepackage{amssymb}  
\usepackage{cite}
\usepackage{amsfonts}
\usepackage{algorithmic}
\usepackage{graphicx}
\usepackage{textcomp}
\usepackage{multirow}
\usepackage{booktabs}
\usepackage{xcolor}
\usepackage{bbding}
\usepackage{pifont}
\usepackage{bm}
\usepackage{threeparttable}

\usepackage{multirow}               
\usepackage{multicol}               
\usepackage{multirow}               
\usepackage{float}                  
\usepackage{makecell}               
\usepackage{booktabs}               
\usepackage{array}
\usepackage{colortbl}  
\usepackage{xcolor}
\usepackage{hyperref}
\definecolor{turquoise}{cmyk}{0.65,0,0.1,0.3}
\definecolor{purple}{rgb}{0.65,0,0.65}
\definecolor{dark_green}{rgb}{0, 0.5, 0}
\definecolor{orange}{rgb}{0.8, 0.6, 0.2}
\definecolor{red}{rgb}{0.8, 0.2, 0.2}
\definecolor{darkred}{rgb}{0.6, 0.1, 0.05}
\definecolor{blueish}{rgb}{0.0, 0.3, .6}
\definecolor{light_gray}{rgb}{0.7, 0.7, .7}
\definecolor{pink}{rgb}{1, 0, 1}
\definecolor{greyblue}{rgb}{0.25, 0.25, 1}
\definecolor{igray}{gray}{.9}
\definecolor{pink}{rgb}{1, 0, 1}
\definecolor{pinkishred}{rgb}{0.9647058823529412, 0.6, 0.8196078431372549}
\definecolor{ForestGreen}{RGB}{21,155,82}

\title{\LARGE \bf
Fully Automated SAM for Single-source Domain Generalization in Medical Image Segmentation
}

\author{Huanli Zhuo, Leilei Ma, Haifeng Zhao\textsuperscript{\ding{41}}, Shiwei Zhou, Dengdi Sun\textsuperscript{\ding{41}}, and Yanping Fu
\thanks{All authors are with the Anhui Provincial Key Laboratory of Multimodal Cognitive Computation, School of Computer Science and Technology, Anhui University, Hefei, P.R. China. Corresponding author: Haifeng Zhao (\href{mailto:senith@163.com}{{\texttt{senith@163.com}}}) and Dengdi Sun (\href{mailto:sundengdi@163.com}{{\texttt{sundengdi@163.com}}}). 
This project is funded by the National Natural Science Foundation of China (61860206004, 62076005, 62472004), Provincial Quality Project of Education in the New Era in 2023 (Postgraduate Education 2023lhpysfjd009), and the University Synergy Innovation Program of Anhui Province, China (GXXT-2021-002,GXXT-2022-029).  
}}

\begin{document}

\maketitle
\thispagestyle{empty}
\pagestyle{empty}

\begin{abstract}
Although SAM-based single-source domain generalization models for medical image segmentation can mitigate the impact of domain shift on the model in cross-domain scenarios, these models still face two major challenges. First, the segmentation of SAM is highly dependent on domain-specific expert-annotated prompts, which prevents SAM from achieving fully automated medical image segmentation and therefore limits its application in clinical settings. Second, providing poor prompts (such as bounding boxes that are too small or too large) to the SAM prompt encoder can mislead SAM into generating incorrect mask results. Therefore, we propose the FA-SAM, a single-source domain generalization framework for medical image segmentation that achieves fully automated SAM. FA-SAM introduces two key innovations: an Auto-prompted Generation Model (AGM) branch equipped with a Shallow Feature Uncertainty Modeling (SUFM) module, and an Image-Prompt Embedding Fusion (IPEF) module integrated into the SAM mask decoder. Specifically, AGM models the uncertainty distribution of shallow features through the SUFM module to generate bounding box prompts for the target domain, enabling fully automated segmentation with SAM. The IPEF module integrates multiscale information from SAM image embeddings and prompt embeddings to capture global and local details of the target object, enabling SAM to mitigate the impact of poor prompts. Extensive experiments on publicly available prostate and fundus vessel datasets validate the effectiveness of FA-SAM and highlight its potential to address the above challenges.
\end{abstract}

\section{INTRODUCTION}
\begin{figure}[t] 
    \centering
    \includegraphics[width=1\linewidth]
    {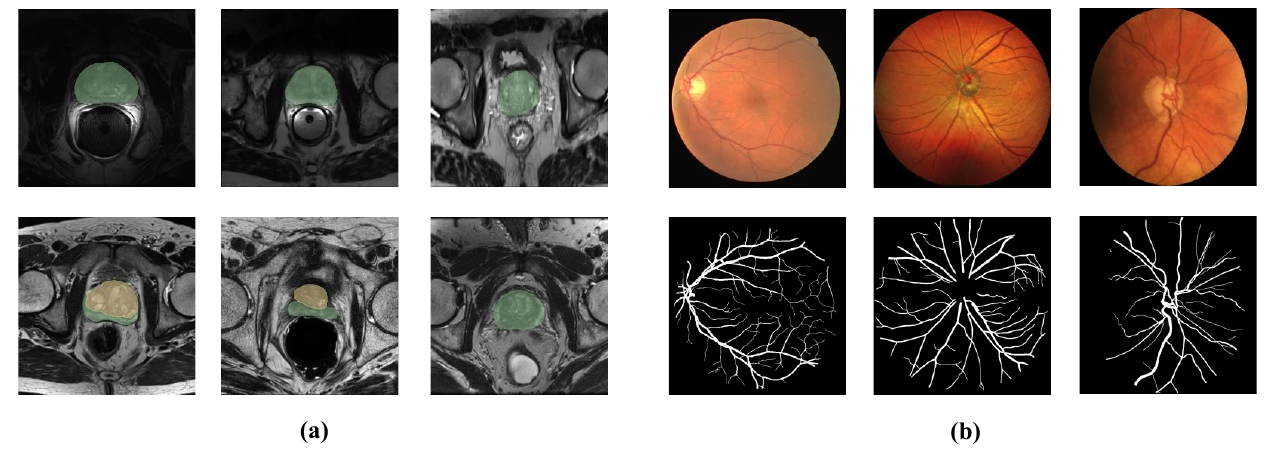}
    \vspace{-2.0em}
    \caption{(a) is prostate MRI collected from six different clinical centers across three public datasets; (b) is retinal fundus vessel images from three public datasets. As can be seen from the figure, there are significant differences in the data from different sites.}
    \vspace{-1.0em}
    \label{fig1}
\end{figure}

Recently, deep neural networks (DNNs) have significantly progressed in various computer vision tasks~\cite{tang2024wfss,ma2025correlative,ma2024text,sun2023uav,ma2023semantic}, including but not limited to medical images~\cite{zhao2024domain,zhou2025111235}. However, these advances typically assume that the training and testing data are independently and identically distributed (i.i.d.)~\cite{sun2023high}. As shown in Fig.\ref{fig1}, this assumption is usually invalid in real-world healthcare settings. Imaging devices, scanning protocols, and other factors lead to distribution differences between images from different medical centers. This difference, known as domain shift, is the primary reason why a model trained on the source domain performs poorly in the target domain. Moreover, due to privacy concerns surrounding medical image data and the high cost of data annotation, it is difficult to obtain large annotated datasets from both the source and target domains. Therefore, Single-source Domain Generalization (SDG) has garnered increasing attention~\cite{zhou2022domain}.

Generally, SDG refers to enhancing model performance in an unseen target domain using only training data from a single source domain. There have been many research findings on SDGs. For example, SLAug~\cite{su2023rethinking} increases global and local diversity of training data to address domain migration issues. EFDM~\cite{zhang2022exact} employs normalization techniques, assuming that feature statistics can capture domain differences. These data-driven SDG methods focus on expanding the training data to model potential feature distributions. Recently, the Segment Anything Model (SAM)~\cite{kirillov2023segment} has been trained on large-scale datasets to create a general-purpose segmentation model, demonstrating impressive performance across the natural image segmentation task~\cite{zhang2023comprehensive,zhu2024weaksam}.
For example, MedSAM~\cite{ma2024segment} fine-tunes SAM on a dataset containing over one million pairs of medical images and masks to enhance its performance in medical image tasks. SAM-U~\cite{deng2023sam} improves segmentation accuracy by using multiple box prompts for better medical image segmentation results. However, SAM faces two major limitations in fields where accuracy is of utmost importance, such as those in SDG medical image segmentation.

\textbf{First, SAM-based segmentation is dependent on manual prompts from medical experts}~\cite{zhang4878606ur,wang2024leveraging}. This means that without expert annotation prompts, SAM cannot achieve fully automated medical image segmentation. Specifically, SAM, which lacks medical knowledge, cannot accurately segment medical images without prompts. Therefore, the performance of SAM relies on manual prompts from medical experts, which is very time-consuming and costly, thus limiting the application of SAM in clinical environments.

\textbf{Second, poor prompts provided to the SAM prompt encoder can mislead SAM into generating incorrect masks}~\cite{gao2024desam,yao2023false}. This means that SAM is more sensitive to poor prompts. Specifically, the Transformer-based SAM mask decoder prioritizes capturing global and low-frequency feature information while neglecting the details of local features. Therefore, when providing SAM with some poor prompts (Such as bounding boxes being too large or too small), the SAM mask decoder fails to fully capture global and local feature information from image embeddings and prompt embeddings, thus unable to alleviate the impact of poor prompts on SAM.

To address these limitations, we propose a novel SAM-based framework, FA-SAM. In this framework, we propose an Auto-prompted Generation Model (AGM) branch, which is an additional trainable segmentation network for single-domain generalization, primarily designed to generate bounding box prompts for the fully automatic segmentation of SAM. Due to the domain shift between the source and target domains, the AGM will inevitably generate some low-quality bounding box prompts for the target domain (\emph{e.g.}, bounding boxes smaller than the masks). Therefore, to minimize the generation of poor-quality prompts, we embed a Shallow Feature Uncertainty Modeling (SUFM) module within the AGM branch. The SUFM module introduces uncertainty into the feature statistics of the shallow convolutional layers of the encoder, aiming to simulate the data distribution of the unseen target domain to alleviate domain shift between the source and target domains. Furthermore, we introduce an Image-Prompt Embedding Fusion (IPEF) module in the SAM mask decoder. This module captures the global features and local details of the target to be segmented by fusing multi-scale information from the image embeddings and prompt embeddings of SAM. By alleviating the impact of poor prompts on SAM through the IPEF module, it enhances the segmentation performance of SAM.

In summary, our contributions consist of the following:
\begin{itemize}
\item We propose the AGM branch, which incorporates the SUFM module. This effectively alleviates domain shift in SDG, enabling the AGM to generate bounding box prompts for the target domain while minimizing the generation of poor-quality prompts. This lays the foundation for fully automated segmentation with SAM.
\item We introduce the IPEF module in the SAM mask decoder. By effectively fusing the embeddings from the SAM image encoder and prompt encoder, it reduces the sensitivity of SAM to poor-quality prompts, enhancing its segmentation performance.
\item The experimental results show that FA-SAM outperforms some existing SDGs. For instance, in the prostate dataset, FA-SAM achieves a 4.99\% improvement in average Dice score over the best-performing SAMMed. 
\end{itemize}

\section{Related Work}
\subsection{Segment Anything Model }
SAM~\cite{kirillov2023segment}, the first segmented everything big model released by Meta AI, demonstrates remarkable proficiency in a wide range of natural image segmentation tasks. Nevertheless, its performance in medical image segmentation is less than optimal, primarily due to its lack of domain-specific knowledge. This is particularly evident in tasks involving boundary blurring or the segmentation of small or complex structures, such as tiny organs~\cite{lin2023samus}. 

In addressing this limitation, several approaches have been proposed, including DeSAM~\cite{gao2024desam}, SkinSAM~\cite{hu2304skinsam}, and MedSAM~\cite{zhang2024segment}. DeSAM modifies the mask decoder of SAM to decouple mask generation and prompt information embedding while utilizing pre-trained weights. However, these methods remain dependent on manual prompts, resulting in semi-automatic segmentation. In contrast, our proposed method achieves fully automated segmentation with SAM.

\subsection{Single-source domain generalization}
Single-source Domain Generalization is a technique that aims to enhance the performance of a model in an unseen target domain when trained on data from only one source domain. The main challenge of SDG lies in the domain shift between the source and target domains. Therefore, common research in SDG mainly focuses on increasing the diversity of training data. For instance, CSDG~\cite{ouyang2022causality} exposes the segmentation model to different domains by synthesizing training samples with domain shifts. 
Maxstyle~\cite{chen2022maxstyle} expands the domain space using additional noise and worst-case combinations. RandConv~\cite{xu2020robust} employs random convolutions for augmentation. MixStyle~\cite{zhou2021domain} combines style information from randomly selected instances across domains. Unlike previous methods, we introduce uncertainty in shallow features to simulate the target domain distribution.

\begin{figure*}[htbp]
\vspace{0.3cm}
\centering  
\includegraphics[width=0.88\linewidth]{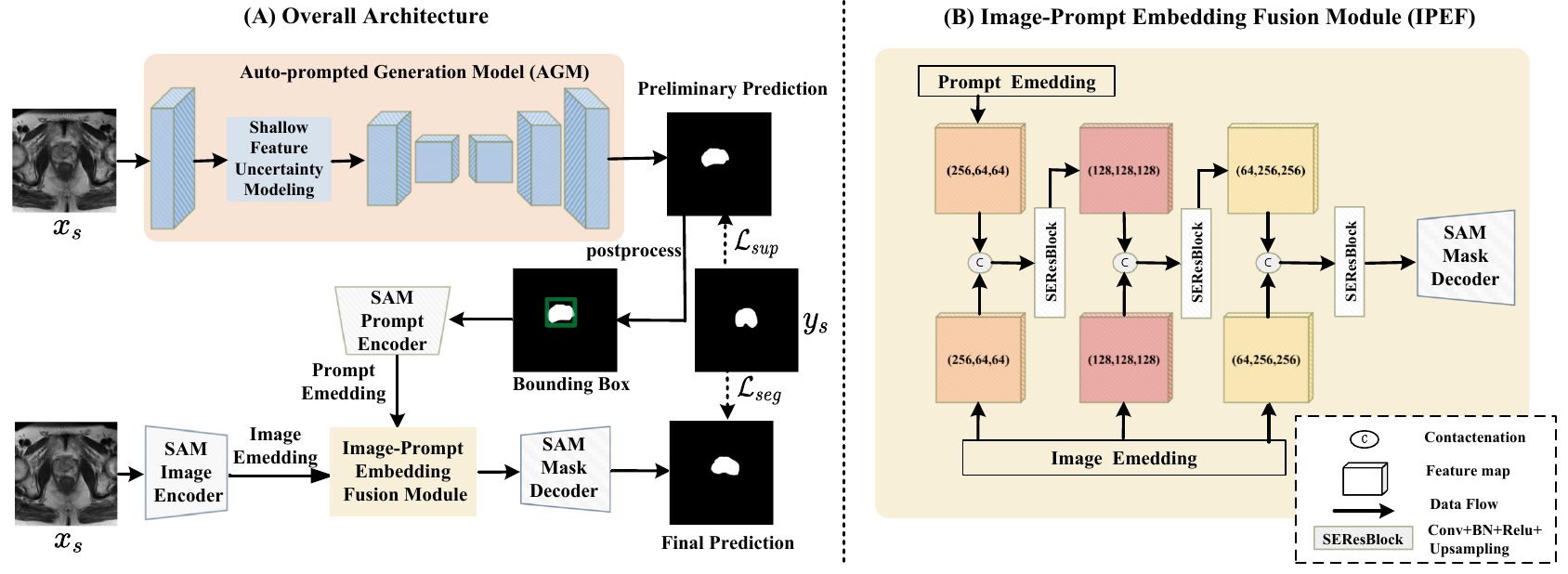}
\caption{(A) The overall architecture of the \textbf{FA-SAM}. The \textbf{AGM} branch with the embedded \textbf{SUFM} module minimizes the generation of poor bounding box prompts for the target domain, enabling fully automatic segmentation of SAM. (B) The \textbf{IPEF} module is introduced into the SAM mask encoder to enhance the ability of SAM to capture global context information and mitigate its sensitivity to poor prompts.}
\label{fig2}
\end{figure*}

\section{METHOD}
We aim to achieve single-domain generalization for fully automated SAM-based medical image segmentation. In the following sections, we will provide a detailed description of the proposed model framework, as shown in Fig.~\ref{fig2}(A). First, we generate the bounding box prompts required by the SAM prompt encoder for the target domain through the AGM branch embedded with the SUFM module, achieving fully automated segmentation with SAM. Second, we introduce the IPEF module into the SAM mask decoder. The IPEF module reduces the sensitivity of SAM to poor prompts by fusing multi-scale information from the SAM image embeddings and prompt embeddings. During these two training processes, we use two losses, $\mathcal{L}_{\text{sup}}$ and $\mathcal{L}_{\text{seg}}$, as the supervision signals.

\subsection{Problem formulation}
Let the train data be sourced from only one domain: $\mathcal{D}_{S}\!\!=\!\!\{({x}_{s}^i, {y}_{s}^i) \!\!\mid\!\! i=1,2, \ldots, {N_s}\}$, where ${x}_{s}^i$ and ${y}_{s}^i$ denote the $i$-th image and its corresponding mask, respectively, and $N_s$ is the number of samples in the source domain. In contrast, the test data originates from unseen domains: $\mathcal{D}_{T}\!\!=\!\!\{\bm{x}_{t}^j \!\mid\! j\!=\!1,2,\ldots, {N_t}\}$, where $N_t$ is the number of target domains. Although $\mathcal{D}_{S}$ and $\mathcal{D}_{T}$ denote the source and target domains, respectively, they share the same label space. For single-domain generalization, the goal is to train a segmentation model with only the source domain $\mathcal{D}_{S}$ visible to accurately segment the image from the invisible target domain $\mathcal{D}_{T}$.

\subsection{Shallow Feature Uncertainty Modeling}
To minimize the generation of poor-quality bounding box prompts for the target domain by the AGM branch, inspired by DSU~\cite{2022Uncertainty}, we embedded a Shallow Feature Uncertainty Modeling (SUFM) module within the AGM. This module is designed to alleviate domain shift between the source and target domains within the AGM branch.

SUFM focuses on introducing uncertainty distributions into the feature statistics of shallow convolutional layers in the encoder. The motivation behind this choice is that the domain shift between the source and target domains primarily involves changes in image features such as color, texture, and contrast~\cite{zhong2019shallow,bui2024meganet}. Shallow convolutional layers are particularly good at capturing these details~\cite{zhong2019shallow}. Furthermore, to better simulate the noise present in real-world medical imaging (\emph{e.g.}, MRI, CT), we combined a noise mechanism based on Poisson and Gaussian
distributions~\cite{muehllehner2006positron}. The introduction of this noise mechanism enables the model to generate diverse feature representations, enhancing the model's uncertainty and thereby improving its generalization ability across different domains.

\begin{figure}[!ht]
    \centering
    \includegraphics[width=1\linewidth]{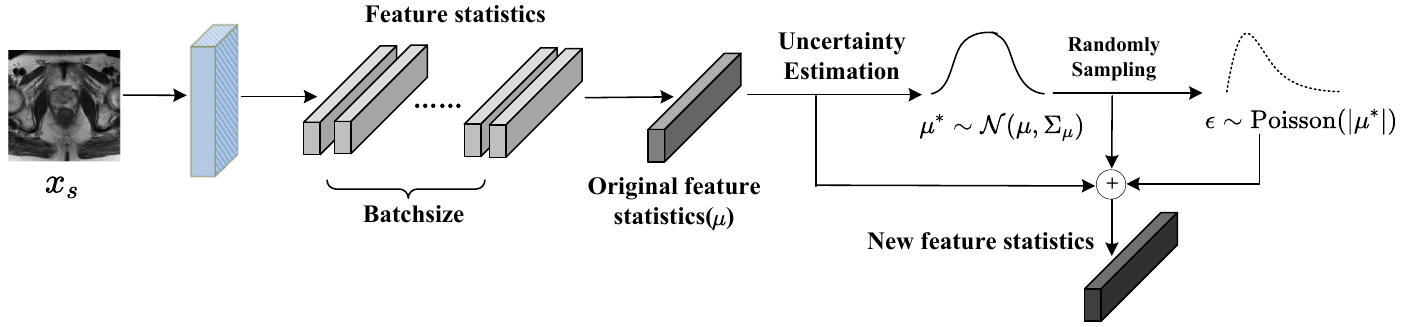}
    \caption{ Structure of our Shallow Feature Uncertainty Modeling (\textbf{SUFM}).}
    \label{fig3}
\end{figure}

To do this, as shown in Fig.\ref{fig3}, the SUFM module generates perturbed new feature statistics by performing uncertainty estimation and random sampling on the original feature statistics, enhancing the robustness of the model to data noise and distributional changes. For the input source image ${x}_{s}$, suppose its shallow encoded features $f \in\mathbb{R}^{B\times C \times H \times W}$. Let $\mu \in \mathbb{R}^{B \times C}$ and $\sigma \in \mathbb{R}^{B \times C}$ represent the channel-wise mean and variance for each instance in the batch, respectively. The following formula can express this:
\begin{equation}
\begin{aligned}
    \mu({x}) &= { {(B\cdot H\cdot W)}^{-1} \cdot}\sum_{b=1}^{B}\sum_{h=1}^{H}\sum_{w=1}^{W} {f}~, \\
    \sigma^2({x}) &= {{{(B\cdot H\cdot W)}^{-1} \cdot}\sum_{b=1}^{B}\sum_{h=1}^{H}\sum_{w=1}^{W} ({f} - \mu({x}))^2}, \label{eq1}
\end{aligned}
\end{equation}
where \textit{B} is the number of samples in a batch, \textit{C} is the channel of the feature map, \textit{H} is the height of the feature map or image, and \textit{W} is the width of the feature map or image. After obtaining the channel-wise mean $\mu$ and variance $\sigma$ of an instance in Eq.~\ref{eq1}, we assume that the distribution of each original feature statistic follows a multivariate Gaussian distribution. The variances of the mean and standard deviation describe the range of uncertainty for different potential feature statistics, that is, the mean and standard deviation of the original feature statistics follow $\mathcal{N}(\mu,\Sigma^2_\mu)$ and $\mathcal{N}(\sigma,\Sigma^2_\sigma)$, respectively. The variances of the mean and standard deviation are expressed as:
\begin{equation}
\begin{aligned}
    \label{eq2}
 {\Sigma^2_{\mu}({x})} &= {{(1/B) \cdot} \sum\nolimits_{i=1}^{B} (\mu({x})-\mathbb{E}_{b}[\mu({x})])^2 }~,\\
    {\Sigma^2_{\sigma}({x})} &= {{(1/B) \cdot} \sum\nolimits_{i=1}^{B} (\sigma({x})-\mathbb{E}_{b}[\sigma({x})])^2 }~,
\end{aligned}   
\end{equation}
where $\Sigma_{\mu}({x}) \in \mathbb{R}^C$ and $\Sigma_{\sigma}({x}) \in \mathbb{R}^C$ represent the uncertainty estimation of the feature mean $\mu$ and feature standard deviation $\sigma$, respectively.

Thus, the uncertainty estimation for the original feature statistics following a multivariate Gaussian distribution is as follows:
\begin{equation}
    \label{eq3}
    \mu^*_\mu \sim \mathcal{N}(\mu,\Sigma^2_\mu),
    \mu^*_\sigma \sim \mathcal{N}(\sigma,\Sigma^2_\sigma).  
\end{equation}

Then, the SUFM module randomly samples the perturbed feature statistics $\mu^*$ from the uncertainty distribution $\mathcal{N}(\cdot,\cdot)$ to simulate the distribution shift of the feature statistics. $\mu^*$ is then further incorporated into the Poisson distribution $\epsilon$ to sample noise, enhancing the feature's adaptability to sparsity or burst disturbances. So, we combine the disturbed statistics $\mu^*$, noise $\epsilon$, and the original feature statistics $\mu$ to generate the new feature statistics mean $\beta({x})$ and standard deviation $\gamma({x})$ as follows:
\begin{equation}
\begin{aligned}
    \beta({x})  & = \mu({x}) + \mu^*_{\mu} \cdot {\Sigma_{\mu}({x})} + \epsilon_\mu ,\\
    \gamma({x}) & = \sigma({x}) + \mu^*_{\sigma} \cdot {\Sigma_{\sigma}({x})} + \epsilon_\sigma.
    \label{eq4}
\end{aligned}
\end{equation}
where $\epsilon_\mu \sim \mathrm{Poisson}(|\mu^*_\mu|)$ and $\epsilon_\sigma \sim \mathrm{Poisson}(|\mu^*_\sigma|)$. Eq.~\ref{eq4} can generate various new feature statistics with different combinations of direction and intensity.

Finally, we use the new feature statistics to replace the original feature statistics for feature transformation, and the final form of the new feature statistics can be expressed as:
\begin{align}   
        f &= \gamma({x}) \cdot {({x} - \mu({x}))} \cdot {\sigma({x})}^{-1} + \beta({x})
\end{align}

\subsection{Auto-prompted Generation Model} 
After generating new feature statistics in the shallow network of AGM, these features continue to enter the convolutional layers, where the loss function $\mathcal{L}_{\text{sup}}$ is used for supervised training of the AGM. Once trained, the AGM generates preliminary predicted masks for both the source and target domains. Then, we use the Breadth-First Search (BFS)~\cite{wang2024leveraging} algorithm to identify the largest connected component in the mask and discard all other components. This post-processing step generates the bounding box prompts needed for the SAM prompt encoder, achieving fully automated segmentation with SAM. This reduces the burden of expert manual annotation and increases the feasibility of SAM in real-world applications. 

Although the AGM branch embedded with the SUFM module minimizes the possibility of generating poor bounding box prompts, the segmentation performance of SAM still relies on some suboptimal prompts~\cite{li2024asps}. To address this issue, we propose the IPEF module, which enhances the segmentation performance of SAM by capturing multi-scale structural information from both the image embeddings and prompt embeddings.

\subsection{Image-Prompt Embedding Fusion Module}
Although the AGM branch has done its best to reduce the likelihood of generating poor bounding box prompts for the target domain, there are inevitably some poor bounding box prompts. Furthermore, due to the lack of progressive upsampling and skip connections in the initial design of the SAM mask decoder, this has become a limitation for preserving high-resolution details in medical images. Therefore, incorporating skip connections is crucial for retaining fine-grained details and effectively distinguishing targets~\cite{he2016deep}. 

Therefore, to mitigate the sensitivity of SAM to poor bounding box prompts, we introduced an Image-Prompt Embedding Fusion (IPEF) module into the SAM mask encoder, as shown in Fig.~\ref{fig2}(B). The IPEF module allows the mask decoder to capture both global features and local details by fusing multi-scale information from the image encoder embeddings and prompt encoder embeddings, thereby alleviating the impact of poor prompts on SAM. Specifically, we first extract image embeddings from the global attention layers of the SAM image encoder ViT-B across different channels (256, 128, 64). These image embeddings, which contain high-dimensional, low-resolution semantic information, are concatenated with the prompt encoder embeddings along the channel dimension. Then, feature enhancement is performed on the concatenated features through Squeeze-and-Excitation Residual (SEResBlocks). Subsequently, features containing low-dimensional, high-resolution semantic information are generated through upsampling and skip connection operations. Finally, the high-resolution features are fed into the SAM mask decoder to produce the final masks for the target domain. As a result, by capturing both global and local information from the image and prompt embeddings, the SAM mask decoder can alleviate the impact of poor prompts on SAM and boost its performance.

\section{Experiments}
\subsection{Datasets}
\textbf{Prostate MRI segmentation.} 
The Prostate Cross-site dataset is a multi-site dataset that comprises prostate T2-weighted MRI data sourced from six distinct data repositories. Specifically, we apply SAML~\cite{liu2020saml} preprocessing to the 2D slices of prostate 3D data. For the source domain (site A), we adopt a split 90\%-10\%, allocating data for training and testing, respectively. Then, we tested the generalizability of our proposed model on five other target domain sites (B-F).

\textbf{Fundus vessel segmentation.} 
Similarly, for fundus segmentation, we used the DRIVE~\cite{2004Ridge} dataset as the source domain (Site A) with a total of 20 paired images and ground truth. To evaluate the generalizability of the proposed method, we used the IOSTAR~\cite{2016Automatic} and LES-AV~\cite{Orlando2018Towards} datasets as target domains (B-C), with 30 and 20 paired images and ground truth, respectively.

\subsection{Implementation Details}
\textbf{Implementation Details.}
We use Meganet~\cite{bui2024meganet} as the backbone network for the AGM branch in the SAM boundary box prompt generation stage. All images are resized to 512$\times$512. During the training process in this stage, the batch size is set to 8, and the model is trained for 200 iterations on the source domain dataset. In the SAM fine-tuning stage, the pre-trained ViT-B from SAM is used, and the image and prompt encoders are frozen during training. All images are resized to 1024$\times$1024. The batch size is set to 4 during training. The SAM fine-tuning stage is trained for 100 iterations. Set the learning rate to 0.001 for all training stages, using the Adam decay optimizer. In addition, all our experiments were conducted on a device equipped with an NVIDIA GeForce RTX 3090 and used a fixed random seed.

\textbf{Training Loss Settings.}
The losses $\mathcal{L}_{\text{sup}}$ and $\mathcal{L}_{\text{seg}}$ used in the two training stages are both a combination of cross-entropy loss and dice loss.

\textbf{Evaluation metric.}
The main metric for evaluating semantic segmentation performance is the Dice coefficient, a set similarity measure commonly used to calculate the similarity between two samples. During the testing phase, it is used to assess the overlap between the predicted results and the ground truth labels.

\subsection{Comparison to State-of-the-Art (SOTA) Results}
\textbf{Performance on the Prostate MRI dataset.} 
We conducted a comprehensive comparison of FA-SAM with SOTA methods on a prostate MRI dataset, as shown in Table \ref{tab1}. Specifically, we compared FA-SAM with data augmentation-based and SAM model-based single-domain generalization segmentation methods. Our method outperformed all others, with FA-SAM showing significant improvements over the best-performing SAMMed~\cite{wang2024leveraging} in domains B and C. The Dice coefficients increased from 77.29\% to 87.00\% and from 73.98\% to 82.93\%, respectively, demonstrating the effectiveness of our approach. Moreover, FA-SAM achieved the best overall performance in all unseen domains, highlighting its strong generalization ability and robustness.

\begin{table}[h]
\renewcommand\arraystretch{1.0}
\caption{Comparison of different segmentation methods using Dice score (\%) for Prostate MRI dataset.}
\vspace{-0.5em}
\label{tab1}
\centering
\resizebox{0.99\linewidth}{!}{
\begin{tabular}{l|cccccc|c}
\toprule
\textbf{Model}& \textbf{A} & \textbf{B} & \textbf{C} & \textbf{D }& \textbf{E} & \textbf{F}&\textbf{Avg} \\
\midrule
RSC~\cite{huang2020self}       & 72.81 & 70.18 & 49.18 & 74.11 & 54.73 & 43.69 & 60.78 \\
AdvBias~\cite{chen2020realistic}   & 78.15 & 62.24 & 54.73 & 72.65 & 53.14 & 51.00 & 61.98 \\
MixStyle~\cite{zhou2021domain}  & 73.24 & 58.06 & 44.75 & 66.78 & 49.81 & 49.73 & 57.06 \\
RandConv~\cite{xu2020robust}        & 77.28 & 60.77 & 53.54 & 66.21 & 52.12 & 36.52 & 57.74 \\
MaxStyle~\cite{chen2022maxstyle}    & 81.52 & 70.27 & 62.09 & 58.18 & 70.04 & 67.77 & 68.27 \\
CSDG~\cite{ouyang2022causality} & 80.72 & 68.00 & 59.78 & 72.40 & 68.67 & 70.78 & 70.06 \\
\midrule
MedSAM~\cite{ma2024segment}       & 72.32 & 73.31 & 61.53 & 64.46 & 68.89 & 61.39 & 66.98 \\
DeSAM~\cite{gao2024desam}        & 82.30 & 78.06 & 66.65 & 82.87 & 77.58 & 79.05	& 77.75 \\
SAMMed~\cite{wang2024leveraging}  & 84.08 & 77.29 & 73.98 & 82.40 & \textbf{80.47}& 79.04 & 79.54  \\  
\midrule
Ours&	 \textbf{88.71}&\textbf{87.00}&	\textbf{82.93}&	\textbf{84.60}&	75.98& \textbf{87.98}& \textbf{84.53}\\ 
\bottomrule
\end{tabular}}
\end{table}
\textbf{Performance on the Fundus vessel dataset.} 
We also evaluated the cross-domain generalization performance of our method on the fundus vascular dataset, with results shown in Table~\ref{tab2}. We compared it with domain adaptation methods, data augmentation-based single-domain generalization methods, and SAM-based single-domain generalization methods. The experimental results indicate that our FA-SAM method outperforms all comparison methods in terms of average performance on this dataset. This highlights the exceptional effectiveness of FA-SAM in addressing the SDG problem in medical image segmentation and demonstrates its strong generalization capability across domains.
\begin{table}[h]
\vspace{0.3cm}
\setlength{\tabcolsep}{4mm}
\renewcommand\arraystretch{0.7}
\caption{Comparison of different segmentation methods using Dice score (\%) for fundus vessel dataset.}
\vspace{-0.5em}
\label{tab2}
\centering
\begin{tabular}{l|cc|c}
\toprule 
\textbf{Model}& \textbf{B }& \textbf{C}&\textbf{Avg} \\
\midrule
CutMix\cite{yun2019cutmix} &64.82&71.48& 68.15\\
DSA\cite{han2021deep} &65.58&71.03&68.30\\
\midrule
BigAug\cite{2021Generalizable}& 62.43&	66.87&64.65\\
{RandConv~\cite{xu2020robust}}&62.01&	67.98&64.99\\
{CSDG~\cite{ouyang2022causality}}&	61.94&	68.30&65.12\\
SLAug\cite{su2023rethinking}&	62.36&	68.79&65.57\\
SAN-SAW\cite{2022Semantic}&63.21&	62.92&63.06\\
FreeSDG\cite{2023Frequency}& 59.72&	63.53&61.62\\
RAS$^{4}$DG~\cite{Qiao_2024_RAS4DG}&65.86&\textbf{72.88}&69.37\\
\midrule
DeSAM~\cite{gao2024desam} &68.32&70.38&69.35\\
SAMMed~\cite{wang2024leveraging} &67.18&68.73&67.95\\
\midrule
Ours	&\textbf{69.29}&71.67&\textbf{70.48}\\
\bottomrule
\end{tabular}
\end{table}

\begin{figure}[htbp]
    \centering
    \includegraphics[width=0.9\linewidth]{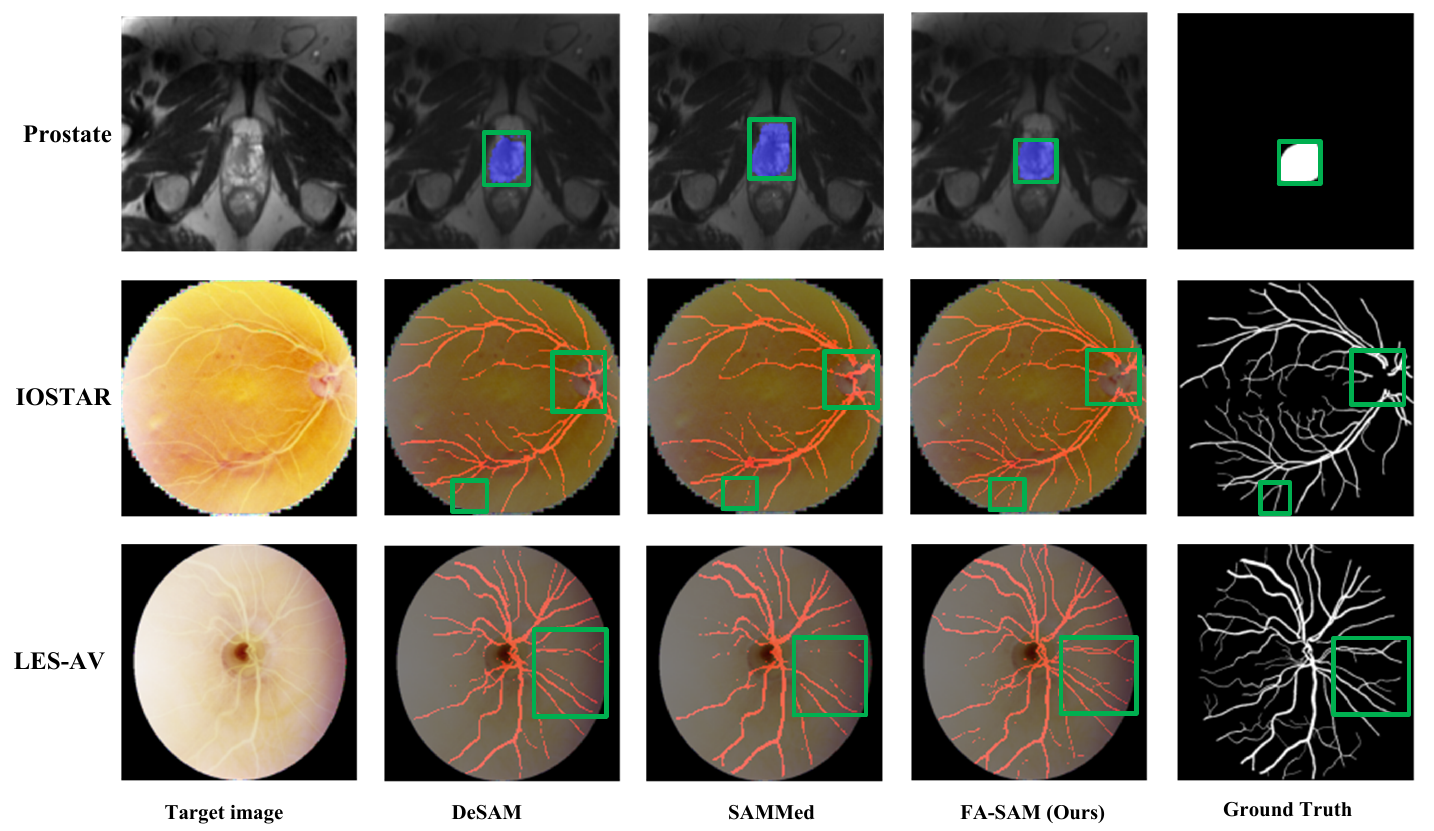}
    \caption{Qualitative comparison of FA-SAM with state-of-the-art methods on the Prostate dataset (site C) and Fundus dataset (site B-C).}
    \label{fig4}
\end{figure}

\textbf{Qualitative Comparison.}
Fig. \ref{fig4} visualizes the segmentation results of various models on unseen domains, including DeSAM~\cite{gao2024desam}, SAMMed~\cite{wang2024leveraging}, FA-SAM (Ours), and Ground Truth (GT). The first row displays segmentation results for a sample from the prostate MRI dataset (unseen domain site C), while the second and third rows show results for two unseen domains (site B and site C) from the fundus vessel dataset. These visualizations offer a clear and intuitive comparison, revealing that our FA-SAM method produces more accurate and complete segmentation shapes, with sharper and more precise boundary delineation.

\begin{table}[htbp]
\setlength{\tabcolsep}{0.7mm}
\renewcommand\arraystretch{1.2}
\caption{Effectiveness of each module using Dice score (\%) for prostate MRI dataset.}
\vspace{-0.5em}
\label{tab3}
\centering
\resizebox{0.99\linewidth}{!}{
\begin{tabular}{cccc|ccccccc}
\toprule
\textbf{SUFM} &\textbf{AGM}$^{*}$&\textbf{IPEF}&\textbf{SAM}& \textbf{A} & \textbf{B} & \textbf{C} & \textbf{D} & \textbf{E} & \textbf{F}&\textbf{Avg}  \\
\midrule
 & \checkmark & &  \checkmark& 88.52&	82.68&	75.02&	86.26&	69.24&	86.11&81.31\\
 \checkmark& \checkmark & & \checkmark&	\textbf{88.80} &86.73&82.62&83.14&	73.43&87.02&83.62\\
 & \checkmark & \checkmark & \checkmark&	{88.65}&{85.06}&	81.72&	\textbf{86.44}&	71.32&	{87.96}&83.52\\
\checkmark& \checkmark &\checkmark & \checkmark & {88.71}&\textbf{87.00}&\textbf{82.93}&{84.60}&	\textbf{75.98}&\textbf{87.98} &\textbf{84.53}\\
\bottomrule
\end{tabular}}
    \begin{tablenotes}
         \footnotesize
         \item[1] The * denotes an AGM branch without the embedded SUFM module.
    \end{tablenotes}
\end{table}
\subsection{Ablation study}
In this section, we explore the impact of each component of our method on the final predictions.
We chose the cross-site prostate dataset for the ablation study.

\textbf{Effectiveness of each module.}
As shown in Table \ref{tab3}, after embedding the SUFM module into the AGM branch, there was a significant improvement in the Dice scores across all target domains. Specifically, for target domains that previously performed poorly, such as Site C and Site E, the Dice coefficients increased by 7.6\% and 4.19\%, respectively. This indicates that the SUFM module effectively mitigated the domain shift issue between the source and target domains, achieving stronger performance on the target domains and enabling the AGM branch to minimize the generation of poor-quality prompts. Furthermore, after introducing the IPEF module into the SAM mask decoder, the Dice scores for all target domains continued to improve. This suggests that the IPEF module enhanced the ability of the SAM mask decoder to effectively capture semantic information. Finally, when the SUFM and IPEF modules were integrated, the Dice scores achieved the best segmentation results, highlighting the complementary contributions of each module in improving performance.

\begin{table}[h]
\renewcommand\arraystretch{0.9}
\caption{Effects of different inserted positions using Dice score (\%) for prostate MRI dataset.}
\vspace{-0.5em}
\label{tab4}
\centering
\resizebox{0.99\linewidth}{!}
{
\begin{tabular}{c|ccccc}
\toprule
\textbf{Inserted Positions}&\textbf{0-1}&\textbf{2}&\textbf{3}&\textbf{4}& \textbf{5}   \\
\midrule
Preliminary Prediction  & \textbf{81.51}  & 76.80 &71.63 &70.68& 69.59	\\
Final Prediction        & \textbf{84.53}  & 80.14 &79.94 &79.48& 77.25	\\
\bottomrule
\end{tabular}}
\end{table}
\textbf{Effects of different inserted positions.} 
To verify the effectiveness of the SUFM module in modeling uncertainty at shallow convolutional layers, we inserted it at different positions within the convolutional layers of the AGM branch for experimentation. The positions where the SUFM module was inserted are labeled as Layer 0 (refers to Convolution Block 0), Max Pooling Layer, and Layers 1, 2, 3, and 4 (refers to Convolution Blocks 1, 2, 3, 4, and 5, respectively). The results are summarized in Table \ref{tab4}, showing that the best performance is achieved when the SUFM module is inserted between Layer 0 and Layer 1 during both the initial and final prediction stages. The experimental results demonstrate the effectiveness of the SUFM module. Placing it in earlier convolutional layers enables the AGM branch to capture target domain details related to potential changes in the source domain, helping to alleviate the domain shift between the source and target domains.

\begin{table}[htbp]
\setlength{\tabcolsep}{1.5mm}
\renewcommand\arraystretch{1}
\caption{Effects of choosing different distributions using Dice score (\%) for prostate MRI dataset.}
\vspace{-0.5em}
\label{tab5}
\centering
\resizebox{0.35\textwidth}{!}{
\begin{tabular}{c|ccc}
\toprule
\textbf{Distribution}&\textbf{Gaussian} &\textbf{Poisson} &\textbf{United} \\
\midrule
Preliminary Prediction& 72.46& 78.93 & \textbf{81.51} 	\\
Final Prediction& 76.96& 82.53 & \textbf{84.53} \\
\bottomrule
\end{tabular}}
\end{table}
\textbf{Effects of choosing different distributions.}
In the DSU~\cite{2022Uncertainty} method, the default configuration uses a Gaussian distribution for uncertainty estimation. However, this approach does not fully account for the Poisson noise in real-world medical imaging data, such as MRI and CT scans~\cite{muehllehner2006positron}. To this end, we introduced a dual disturbance mechanism in the SUFM module, where Gaussian distribution sampling simulates the continuous variation of feature statistics, and Poisson noise injects discrete disturbances to enhance feature diversity. In this work, we inserted the SUFM module into the 0-1 convolutional layer and compared the performance of Gaussian and Poisson distributions in Table \ref{tab5}. The results show that the mechanism integrating Gaussian and Poisson distributions improves the model's overall uncertainty and strengthens its generalization ability.

\begin{table}[t]
\vspace{0.3cm}
\renewcommand\arraystretch{1.15}  
\caption{Effects of the IPEF module using Dice score (\%) for prostate MRI dataset.}
\vspace{-0.5em}
\label{tab6}
\centering
\resizebox{1.0\linewidth}{!}
{
\begin{tabular}{c|cccccc|c}
\toprule
\textbf{Model} & \textbf{A} & \textbf{B} & \textbf{C} & \textbf{D} & \textbf{E} & \textbf{F}&\textbf{Avg}  \\
\midrule
SAM+GT&92.79&90.71&91.35&91.31&87.64&89.39&90.53\\
SAM+IPEF+GT&93.18&91.77&91.42&92.01&89.75&	91.13&91.54\\
FA-SAM(Ours) &88.71&87.00&	82.93&	84.60&	75.98&	87.98&84.53\\ 
\bottomrule
\end{tabular}}
\end{table}
\textbf{Effects of the IPEF module.}
To assess the effectiveness of the IPEF module, we conducted two experiments: one with the IPEF module and one without. In both experiments, SAM was fine-tuned using Ground Truth (GT) prompts, and the results are presented in Table \ref{tab6}. Although our method still lags behind SAM using Ground Truth (GT) as prompts, the approach combining SAM with GT prompts and the IPEF module achieves a Dice score that is 1.01\% higher than the method without the IPEF module. This improvement highlights the effectiveness of the IPEF module in fine-tuning SAM and demonstrates its potential to enhance SAM's ability to adapt to unknown domains.

\begin{figure}[htbp]
    \centering
    \includegraphics[width=0.9\linewidth]{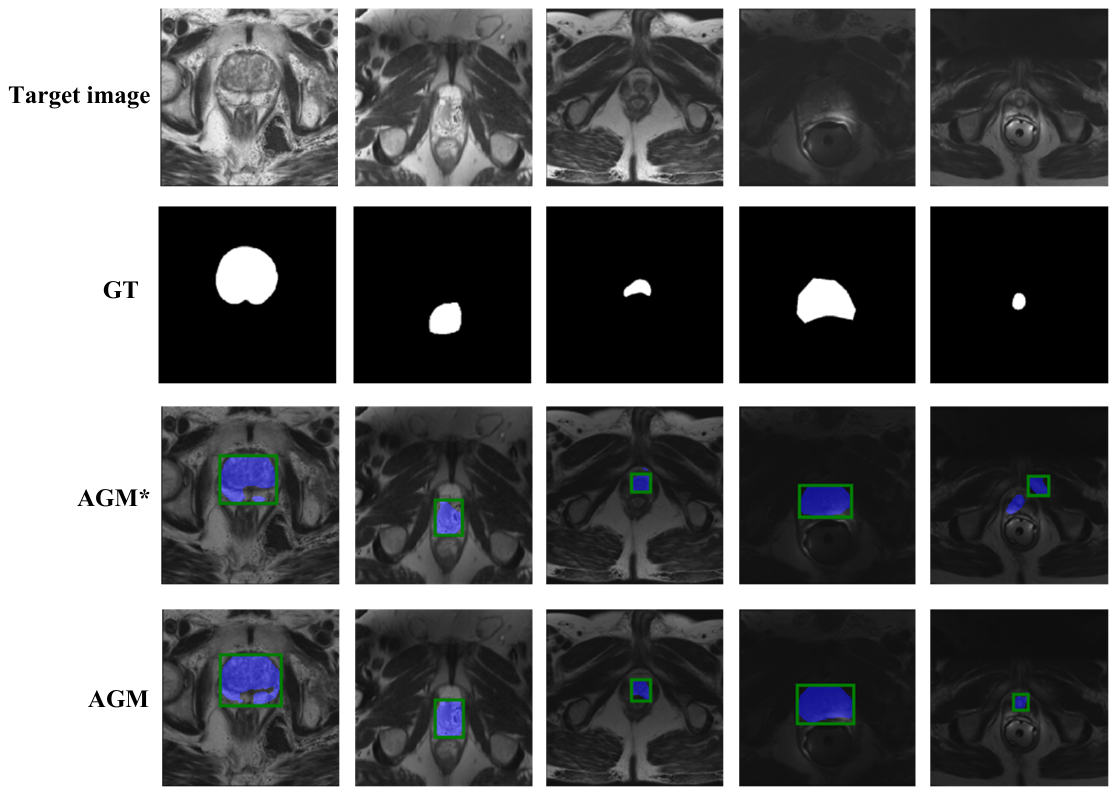}
    \caption{ This visualization result shows the AGM branch generating SAM boundary box prompts. The target image comes from site B-F in the prostate dataset. The * indicates an AGM branch without the embedded SUFM.}  
    \label{fig5}
\end{figure}

\textbf{Visualization.} 
In Fig.\ref{fig5}, we evaluate the effectiveness of the AGM branch embedded with the SUFM through visual results on the prostate target domain. The visualization results show that the bounding box prompts generated by the AGM branch with the embedded SUFM are more accurate.

\section{CONCLUSION}
This paper proposes a single-domain generalization framework for medical images in fully automated SAM. The framework introduces the AGM branch, which is embedded with an SUFM module. This branch models the uncertainty of shallow feature statistics generated by convolutional layers during the training process, minimizing the generation of poor-quality prompts for the target domain and achieving fully automated segmentation with SAM. Additionally, the method incorporates the IPEF module into the SAM mask encoder, which alleviates the sensitivity of the SAM mask encoder to poor prompts by fusing multi-scale features from image encoder embeddings and prompt encoder embeddings. Experimental results on the fundus and prostate datasets demonstrate the effectiveness of this method, achieving fully automated segmentation with SAM.

\section*{ACKNOWLEDGMENT}
{\small 
We sincerely thank Professor \href{https://scholar.google.com/citations?user=0qaDapcAAAAJ&hl=en}{Bin Luo} for his generous support of this project.
We are grateful to Yufei Zhang, Shuo Xu, Xu Tang, Haiyang Li, Wenhai Qin, and Wanzhen Hou from Anhui University for their valuable discussions.
We also thank the reviewers for their comments and suggestions.
Finally, we acknowledge the High-performance Computing Platform of Anhui University for providing computational resources for this project.
}

\bibliographystyle{IEEEtran}  
\bibliography{IEEEabrv,ref}


\end{document}